\documentclass{article}
\usepackage{iclr2026_conference,times}

\usepackage{graphicx}
\usepackage{booktabs}
\usepackage{amsmath}
\usepackage{amssymb}
\usepackage{siunitx}
\usepackage{xcolor}
\usepackage{colortbl}
\usepackage{multirow}
\usepackage{hyperref}
\usepackage{url}
\usepackage{natbib}

\definecolor{lightgray}{RGB}{245,245,245}

\newcommand{\secondbest}[1]{\underline{#1}}

\usepackage{amsmath,amsfonts,bm}

\def\eqref#1{equation~\ref{#1}}

\def\1{\bm{1}}

\DeclareMathAlphabet{\mathsfit}{\encodingdefault}{\sfdefault}{m}{sl}
\SetMathAlphabet{\mathsfit}{bold}{\encodingdefault}{\sfdefault}{bx}{n}

\title{\raggedright Infinity and Beyond: Compositional Alignment in VAR and Diffusion T2I Models}

\iclrfinalcopy 

\author{%
Hossein Shahabadi${}^{1}$ \, Niki Sepasian${}^{1}$ \, Arash Marioriyad${}^{1}$\\
\bf{Ali Sharifi-Zarchi${}^{1}$ \, Mahdieh Soleymani Baghshah${}^{1}$} \\
${}^{1}$Sharif University of Technology
}

\begin{document}

\maketitle

\begin{abstract}

Achieving compositional alignment between textual descriptions and generated images - covering objects, attributes, and spatial relationships - remains a core challenge for modern text-to-image (T2I) models. Although diffusion-based architectures have been widely studied, the compositional behavior of emerging Visual Autoregressive (VAR) models is still largely unexamined. We benchmark six diverse T2I systems - SDXL, PixArt-$\alpha$, Flux-Dev, Flux-Schnell, Infinity-2B, and Infinity-8B - across the full T2I-CompBench++ and GenEval suites, evaluating alignment in color and attribute binding, spatial relations, numeracy, and complex multi-object prompts. Across both benchmarks, Infinity-8B achieves the strongest overall compositional alignment, while Infinity-2B also matches or exceeds larger diffusion models in several categories, highlighting favorable efficiency–performance trade-offs. In contrast, SDXL and PixArt-$\alpha$ show persistent weaknesses in attribute-sensitive and spatial tasks. These results provide the first systematic comparison of VAR and diffusion approaches to compositional alignment and establish unified baselines for the future development of the T2I model.
\end{abstract}

\section{Introduction}


Recent advances in text-to-image (T2I) generation have enabled models to synthesize highly realistic, semantically rich, and stylistically diverse images at scale \citep{ramesh2021zero, ramesh2022hierarchical, saharia2022photorealistic, podell2023sdxl}. However, despite their impressive perceptual quality, contemporary T2I models continue to exhibit substantial limitations in compositional alignment, the ability to faithfully bind objects, attributes, and spatial or non-spatial relations described in natural language into coherent visual outputs. A growing body of work demonstrates that strong visual fidelity does not equate to reliable compositional correctness, particularly in multi-object or attribute-heavy prompts, where models frequently violate attribute bindings, confuse spatial configurations, or hallucinate unintended elements \citep{huang2025t2i, ghosh2023geneval, chefer2023attend}.



Although prior work has examined compositionality in diffusion-based T2I models, the compositional alignment abilities of emerging visual autoregressive (VAR) architectures remain largely unexplored \citep{tian2024visual, han2025infinity}. These models exhibit strong global coherence and semantic richness, yet their capacity to preserve structured multi-entity relationships has not been systematically evaluated. To fill this gap, we assess six modern T2I models - Flux-Dev and Flux-Schnell \citep{flux2024}, SDXL, PixArt-$\alpha$ \citep{chen2023pixart}, Infinity-2B, and Infinity-8B \citep{han2025infinity} - using T2I-CompBench++ \citep{huang2025t2i} as our primary benchmark, and further validate consistency through the full GenEval \citep{ghosh2023geneval} suite.

Across both T2I-CompBench++ and GenEval, Infinity-8B achieves the strongest overall compositional alignment, leading in color, texture, shape, spatial relations, and multi-constraint prompts, while Infinity-2B also surpasses or matches diffusion models in several dimensions despite its smaller scale. Flux-Dev and Flux-Schnell exhibit competitive performance but trail Infinity-8B across most categories, whereas SDXL and PixArt-$\alpha$  consistently underperform, particularly on attribute binding tasks. These results provide a clear quantitative characterization of compositional alignment across diffusion, diffusion-transformer, and VAR-based architectures, establishing Infinity models as strong contenders in T2I generation.

\section{Benchmark Setup}

\subsection{T2I-CompBench++}
We adopt T2I-CompBench++ as our primary evaluation suite \citep{huang2025t2i}. The benchmark covers eight major aspects of text-image compositional alignment: color, texture, and shape binding; non-spatial relations; 2D and 3D spatial relations; counting; and complex multi-attribute compositions. 
Color, texture, and shape measure attribute binding; non-spatial relations capture interactions such as ``holding'' or ``wearing''; 2D and 3D spatial categories evaluate positional and depth relations; numeracy tests correct object counts; and complex compositions assess multi-attribute, multi-object prompts. For each prompt in the dataset, every model generates a single image, and accuracy is computed using the official evaluation pipeline. These metrics collectively capture a wide spectrum of compositional failures, from incorrect attribute assignment to violations of spatial configurations.

\subsection{GenEval}
We include GenEval \citep{ghosh2023geneval} alongside T2I-CompBench++ because the two benchmarks capture complementary aspects of compositional alignment: T2I-CompBench++ relies on detector-driven verification, whereas GenEval uses rule-based constraints, enabling a more holistic evaluation. To validate our findings under an independent evaluation protocol, we also run all models on the full GenEval benchmark. GenEval provides seven metrics - Overall, Single Object, Two Objects, Counting, Colors, Position, and Color Attributes - covering object presence, attribute correctness, and simple spatial reasoning. 
Single Object and Two Objects test object presence; Colors and Color Attributes evaluate correct attribute assignment; Position measures simple spatial layout; Counting evaluates numerical correctness; and the Overall score aggregates these metrics into a single alignment measure.
Unlike T2I-CompBench++, GenEval relies on prompt-structured rules rather than detector-heavy pipelines, making it a complementary sanity check.

\subsection{Evaluation Protocol}


For T2I-CompBench++, we assess each model using four independent random seeds and report seed-averaged results, with full mean and standard deviation values provided in Appendix~\ref{appendix:variance}. For GenEval, we follow the official protocol by generating four samples per prompt; because GenEval aggregates these into a single metric without exposing sample-level variance, we report only the aggregated scores.

\section{Text-to-Image Models}

As described in Table \ref{tab:model_metadata}, our evaluation includes six state-of-the-art open models spanning diffusion, diffusion–transformer hybrid, and autoregressive (VAR-based) architectures. 

\paragraph{SDXL.} 
Stable Diffusion XL is a diffusion model with a large U-Net backbone and CLIP-based text conditioning \citep{podell2023sdxl}. It is widely used due to its high aesthetic quality, making it an important reference point for assessing compositional behavior.

\paragraph{PixArt-$\alpha$.}
PixArt-$\alpha$ is a transformer-based diffusion model for high-resolution image synthesis \citep{chen2023pixart}. Despite its strong visual quality, prior work notes difficulties in object binding and spatial consistency, motivating its inclusion as a contrasting baseline \citep{balaji2022ediff}.

\paragraph{Flux-Dev and Flux-Schnell.} 
Flux models are diffusion-based generators that incorporate transformer blocks and efficient attention mechanisms \citep{flux2024}. Flux-Dev provides higher-quality outputs, while Flux-Schnell is optimized for speed. Their differing configurations allow us to compare quality–speed trade-offs in compositional tasks.

\paragraph{Infinity-2B and Infinity-8B.}
Infinity models adopt a bit-wise visual autoregressive paradigm that predicts hierarchical latent codes across scales via a next-scale autoregressive transformer \citep{han2025infinity}, fundamentally differing from diffusion-based denoising. Infinity-2B and Infinity-8B correspond to mid-sized and large configurations within this framework.

\begin{table}[htbp]
\centering
\caption{Metadata for all T2I models included in our evaluation.}
\label{tab:model_metadata}
\resizebox{\textwidth}{!}{
\begin{tabular}{lcccc}
\toprule
\textbf{Model} &
\textbf{Architecture Type} &
\textbf{Release} &
\textbf{Parameters} &
\textbf{Text Encoder} \\
\midrule
SDXL             & Diffusion (UNet) & 2023 & 3.5B & OpenCLIP ViT-G \\
PixArt-$\alpha$ & Diffusion (Transformer) & 2023 & 0.6B & T5-XXL \\
Flux-Dev     & DiT    & 2024 & 12B  & CLIP, Flan-T5-XXL \\
Flux-Schnell  & DiT     & 2024 & 12B  & CLIP, Flan-T5-XXL \\
Infinity-2B   & VAR   & 2024 & 2B   & Flan-T5-XL \\
Infinity-8B    & VAR  & 2025 & 8B   & Flan-T5-XL \\
\bottomrule
\end{tabular}
}
\end{table}

\section{Results}

\subsection{T2I-CompBench++}
Table~\ref{tab:compbench_main} summarizes category-wise performance on T2I-CompBench++. Infinity-8B emerges as the strongest model, ranking first in seven of the eight evaluation categories and placing a close second in numeracy. Infinity-2B also performs well, forming a middle tier together with the Flux models and consistently outperforming SDXL and PixArt-$\alpha$. 

Among non-VAR models, Flux-Schnell achieves the most balanced results and secures several second-place rankings, while Flux-Dev attains the highest numeracy score. In contrast, SDXL and PixArt-$\alpha$ show limited compositional alignment, particularly in spatial and complex categories, where they consistently occupy the bottom positions.

Overall, our results suggest that VAR-based models perform competitively with diffusion models within the scope of the compositional benchmarks we evaluate, and that sufficiently scaled autoregressive architectures, as demonstrated by Infinity-8B, achieve leading performance across a broad range of tasks.

\begin{table}[htbp]
\centering
\caption{T2I-CompBench++ compositionality scores (higher is better). Best values in \textbf{bold}, second best \secondbest{underlined}. Results are averaged over four runs; additional statistics appear in Appendix~\ref{appendix:variance}.}
\label{tab:compbench_main}
\resizebox{\textwidth}{!}{%
\begin{tabular}{lcccccccc|c}
\toprule
\textbf{Model} &
\textbf{Color} &
\textbf{Texture} &
\textbf{Shape} &
\textbf{Non-Spatial} &
\textbf{2D Spatial} &
\textbf{3D Spatial} &
\textbf{Numeracy} &
\textbf{Complex} &
\textbf{Mean} \\
\midrule
SDXL            & 0.593 & 0.519 & 0.466 & 0.311 & 0.215 & 0.341 & 0.504 & 0.319 & 0.409 \\
PixArt-$\alpha$ & 0.407 & 0.444 & 0.367 & 0.308 & 0.202 & 0.350 & 0.506 & 0.324 & 0.364 \\
Flux-Dev        & \secondbest{0.746} & 0.644 & 0.482 & 0.309 & \secondbest{0.273} & 0.393 & \textbf{0.613} & 0.363 & 0.478 \\
Flux-Schnell    & 0.725 & \secondbest{0.683} & \secondbest{0.559} & \secondbest{0.312} & 0.271 & 0.373 & 0.604 & 0.364 & \secondbest{0.486} \\
Infinity-2B     & 0.741 & 0.636 & 0.480 & 0.310 & 0.240 & \secondbest{0.406} & 0.573 & \secondbest{0.382} & 0.471 \\
Infinity-8B     & \textbf{0.827} & \textbf{0.753} & \textbf{0.604} & \textbf{0.316} & \textbf{0.365} & \textbf{0.414} & \secondbest{0.612} & \textbf{0.397} & \textbf{0.536} \\
\bottomrule
\end{tabular}}
\end{table}

\subsection{GenEval}
GenEval results, shown in Table~\ref{tab:geneval_main}, exhibit the same overall trends observed in T2I-CompBench++. Infinity-8B is the strongest model, achieving the highest Overall score and leading every individual metric. Infinity-2B and Flux-Schnell form a competitive second tier, consistently outperforming SDXL and PixArt-$\alpha$ across most categories. SDXL and PixArt-$\alpha$ again show the weakest compositional alignment, with pronounced errors in positional reasoning and attribute binding.

The consistency between GenEval and T2I-CompBench++ indicates that the performance trends are robust across distinct datasets, despite the benchmarks employing different scoring methodologies.

For additional context, Appendix~\ref{appendix:geneval_extended} compares our results against GenEval scores reported in prior work, including proprietary models (e.g., DALL·E) and test-time optimization methods such as ReNO \citep{eyring2024reno, ramesh2021zero}.

\begin{table}[htbp]
\centering
\caption{GenEval scores. Best values in \textbf{bold}, second best \secondbest{underlined}.}
\label{tab:geneval_main}
\resizebox{\textwidth}{!}{%
\begin{tabular}{lcccccc|c}
\toprule
\textbf{Model} &
\textbf{Colors} &
\textbf{Color Attr.} &
\textbf{Position} &
\textbf{Single Obj.} &
\textbf{Two Obj.} &
\textbf{Counting} &
\textbf{Overall} \\
\midrule
SDXL            
& \secondbest{0.862} & 0.210 & 0.105 & 0.984 & 0.664 & 0.409 & 0.539 \\
PixArt-$\alpha$ 
& 0.801 & 0.093 & 0.068 & 0.978 & 0.505 & 0.438 & 0.480 \\
Flux-Dev        
& 0.766 & 0.470 & 0.185 & 0.988 & 0.785 & \secondbest{0.716} & 0.652 \\
Flux-Schnell    
& 0.785 & 0.505 & 0.263 & \textbf{1.000} & \secondbest{0.894} & 0.597 & 0.674 \\
Infinity-2B     
& 0.830 & \secondbest{0.590} & \secondbest{0.270} & 0.997 & 0.798 & 0.597 & \secondbest{0.680} \\
Infinity-8B     
& \textbf{0.886} & \textbf{0.765} & \textbf{0.578} & \textbf{1.000} & \textbf{0.937} & \textbf{0.778} & \textbf{0.824} \\
\bottomrule
\end{tabular}}
\end{table}

\subsection{Runtime and Memory Observations}

Although computational efficiency is not the primary focus of this study, we report a limited set of runtime and memory measurements (see Appendix~\ref{appendix:runtime}) to contextualize the practical behavior of the evaluated models. The non-VAR models exhibit substantial variation in efficiency: Flux-Schnell is the fastest, whereas Flux-Dev incurs the highest memory usage and slowest generation times; PixArt-$\alpha$ also displays relatively slow inference despite its moderate parameter count. In contrast, Infinity-2B achieves a favorable efficiency–performance trade-off, requiring considerably less GPU memory than the large diffusion models and generating images faster than all diffusion baselines. Infinity-8B necessitated KV-caching to fit on a single A100 GPU, consistent with prior findings that KV-cache usage affects memory consumption and decoding speed but does not alter model outputs or evaluation performance \citep{li2025memory}. As such, its memory profile is not directly comparable to the other models.

\section{Limitations}

Our evaluation covers only two benchmarks, and additional datasets or human studies may reveal complementary aspects of compositional behavior. Moreover, runtime and memory assessments are limited to a single hardware setup and all results rely on automatic evaluators, which can be sensitive to prompt ambiguity and may miss nuanced visual semantics.

\section{Discussion}

Our evaluation reveals clear and consistent patterns in the compositional behavior of contemporary T2I models. Within the VAR family, scale plays a decisive role: Infinity-8B achieves the strongest compositional alignment across both T2I-CompBench++ and GenEval, while the much smaller Infinity-2B remains competitive with strong diffusion baselines. These results indicate that autoregressive architectures, when sufficiently scaled, can deliver robust compositional generation, and that even mid-sized VAR models can rival or surpass considerably larger diffusion models, possibly due to the next-scale autoregressive generation used by VAR models, which explicitly conditions each stage on previously generated visual structure, in contrast to denoising-based generation where global consistency must emerge implicitly over many refinement steps; however, several compositional categories show limited gains with increased scale, indicating that further improvements in these dimensions likely require factors beyond model scaling alone.

In contrast, diffusion and hybrid models exhibit greater variability, with Flux-Schnell and Flux-Dev forming a reliable mid-tier, while SDXL and PixArt-$\alpha$ consistently underperform across most compositional categories. This reinforces that high aesthetic fidelity does not necessarily correlate with accurate multi-object binding, spatial structure, or attribute consistency.

The strong agreement between T2I-CompBench++ and GenEval, despite differing verification methodologies, suggests that the observed trends are stable across evaluation protocols. Both benchmarks identify persistent challenges shared by all models, including spatial reasoning, fine-grained attribute binding, and multi-object scene composition.

Overall, our study provides a unified and systematic comparison of compositional capability across diffusion, hybrid, and visual autoregressive architectures. By establishing consistent baselines and highlighting the roles of scale, training objectives, and model design, we aim to support future work on T2I systems that more reliably encode structured scene information.

\newpage

\bibliography{iclr2026_conference}
\bibliographystyle{iclr2026_conference}

\newpage
\appendix

\section{Additional Variance Measurements}
\label{appendix:variance}

The following tables report the mean and standard deviation across four seeds for all T2I-CompBench++ categories. The main paper reports only mean values for readability.

\begin{table}[htbp]
\centering
\small
\caption{T2I-CompBench++ mean $\pm$ std for basic attribute categories.}
\label{tab:appendix_basic}
\begin{tabular}{lcccc}
\toprule
\textbf{Model} &
\textbf{Color} &
\textbf{Texture} &
\textbf{Shape} &
\textbf{Non-Spatial} \\
\midrule
SDXL             & 0.593 $\pm$ 0.031 & 0.519 $\pm$ 0.013 & 0.466 $\pm$ 0.020 & 0.311 $\pm$ 0.002 \\
PixArt-$\alpha$  & 0.407 $\pm$ 0.028 & 0.444 $\pm$ 0.014 & 0.367 $\pm$ 0.012 & 0.308 $\pm$ 0.001 \\
Flux-Dev         & 0.746 $\pm$ 0.020 & 0.644 $\pm$ 0.016 & 0.482 $\pm$ 0.013 & 0.309 $\pm$ 0.000 \\
Flux-Schnell     & 0.725 $\pm$ 0.042 & 0.683 $\pm$ 0.027 & 0.559 $\pm$ 0.014 & 0.312 $\pm$ 0.002 \\
Infinity-2B      & 0.741 $\pm$ 0.007 & 0.636 $\pm$ 0.009 & 0.480 $\pm$ 0.007 & 0.310 $\pm$ 0.001 \\
Infinity-8B      & 0.827 $\pm$ 0.003 & 0.753 $\pm$ 0.002 & 0.604 $\pm$ 0.004 & 0.316 $\pm$ 0.001 \\
\bottomrule
\end{tabular}
\end{table}

\begin{table}[htbp]
\centering
\small
\caption{T2I-CompBench++ mean $\pm$ std for spatial and composition categories.}
\label{tab:appendix_spatial}
\begin{tabular}{lcccc}
\toprule
\textbf{Model} &
\textbf{2D Spatial} &
\textbf{3D Spatial} &
\textbf{Numeracy} &
\textbf{Complex} \\
\midrule
SDXL             & 0.215 $\pm$ 0.012 & 0.341 $\pm$ 0.007 & 0.504 $\pm$ 0.017 & 0.319 $\pm$ 0.005 \\
PixArt-$\alpha$  & 0.202 $\pm$ 0.012 & 0.350 $\pm$ 0.005 & 0.506 $\pm$ 0.008 & 0.324 $\pm$ 0.003 \\
Flux-Dev         & 0.273 $\pm$ 0.007 & 0.393 $\pm$ 0.015 & 0.613 $\pm$ 0.024 & 0.363 $\pm$ 0.001 \\
Flux-Schnell     & 0.271 $\pm$ 0.016 & 0.373 $\pm$ 0.019 & 0.604 $\pm$ 0.008 & 0.364 $\pm$ 0.012 \\
Infinity-2B      & 0.240 $\pm$ 0.008 & 0.406 $\pm$ 0.003 & 0.573 $\pm$ 0.007 & 0.382 $\pm$ 0.003 \\
Infinity-8B      & 0.365 $\pm$ 0.010 & 0.414 $\pm$ 0.008 & 0.612 $\pm$ 0.006 & 0.397 $\pm$ 0.002 \\
\bottomrule
\end{tabular}
\end{table}

\section{Runtime and Memory Measurements}
\label{appendix:runtime}
Table~\ref{tab:runtime_memory} reports runtime and memory measurements collected using 50 randomly selected prompts from T2I-CompBench++, with all models evaluated on the same prompt set for comparability. Reported memory values correspond to PyTorch's \textit{allocated} CUDA memory, which reflects model-controlled GPU usage. System-level tools such as \texttt{nvidia-smi} may show higher totals due to additional runtime and driver allocations. All models were measured using the same
procedure for consistency.

\begin{table}[htbp]
\centering
\caption{Runtime and memory measurements averaged over 50 prompts.}
\label{tab:runtime_memory}
\begin{tabular}{lccc}
\toprule
\textbf{Model} &
\textbf{Load Mem. (GB)} &
\textbf{Peak Mem. (GB)} &
\textbf{Avg. Time (s)} \\
\midrule
SDXL            & 6.58  & 10.47 & 4.08 \\
PixArt-$\alpha$ & 12.07 & 14.46 & 5.82 \\
Flux-Dev        & 33.32 & 35.71 & 9.86 \\
Flux-Schnell    & 33.29 & 35.68 & 2.74 \\
Infinity-2B     & 7.37  & 15.75 & 2.13 \\
Infinity-8B     & --    & --    & --   \\
\bottomrule
\end{tabular}
\end{table}

\section{Extended GenEval Comparison with Prior Work}
\label{appendix:geneval_extended}

To place our results in the context of prior large-scale and proprietary
text-to-image systems, we additionally report GenEval scores for a range of
models from the literature, including ReNO-enhanced diffusion models and
proprietary systems such as DALL·E. All scores in
Table~\ref{tab:geneval_extended} are taken from the corresponding papers and were
not re-evaluated by us. Scores reported below the midrule are intended as reference points from prior work rather than strictly controlled comparisons, as they may differ in prompt sets, inference settings, and evaluation protocols.

This extended comparison highlights two complementary trends. First, test-time
optimization methods such as ReNO can substantially improve the compositional
alignment of diffusion-based models, with particularly large gains in counting
and attribute-sensitive categories (e.g., SDXL-Turbo, HyperSDXL, and
FLUX-Schnell). Second, despite the absence of any test-time optimization, the
base Infinity-8B model achieves comparable or stronger performance across most
GenEval categories. This suggests that architectural design and model scale play
a substantial role in compositional alignment, even without inference-time
optimization.

We report prior Flux-Schnell results as published and do not attempt to reconcile
numerical differences with our own evaluation, as these differences likely arise
from variations in inference settings and evaluation protocols.

\begin{table}[htbp]
\centering
\caption{Extended GenEval scores from prior work. Rows highlighted in gray correspond to models using test-time optimization (e.g., ReNO). Models above the first midrule are evaluated in this work; models below are reported from external sources. Best values per column are shown in \textbf{bold}.}
\label{tab:geneval_extended}
\resizebox{\textwidth}{!}{%
\begin{tabular}{lcccccc|c}
\toprule
\textbf{Model} &
\textbf{Colors} &
\textbf{Color Attr.} &
\textbf{Position} &
\textbf{Single Obj.} &
\textbf{Two Obj.} &
\textbf{Counting} &
\textbf{Overall} \\
\midrule
SDXL            
& 0.862 & 0.210 & 0.105 & 0.984 & 0.664 & 0.409 & 0.539 \\
PixArt-$\alpha$ 
& 0.801 & 0.093 & 0.068 & 0.978 & 0.505 & 0.438 & 0.480 \\
Flux-Dev        
& 0.766 & 0.470 & 0.185 & 0.988 & 0.785 & 0.716 & 0.652 \\
Flux-Schnell    
& 0.785 & 0.505 & 0.263 & \textbf{1.000} & 0.894 & 0.597 & 0.674 \\
Infinity-2B     
& 0.830 & 0.590 & 0.270 & 0.997 & 0.798 & 0.597 & 0.680 \\
Infinity-8B     
& 0.886 & \textbf{0.765} & \textbf{0.578} & \textbf{1.000} & \textbf{0.937} & 0.778 & \textbf{0.824} \\
\midrule
DALL-E\,2       
& 0.770 & 0.190 & 0.100 & 0.940 & 0.660 & 0.490 & 0.520 \\
DALL-E\,3       
& 0.830 & 0.450 & 0.430 & 0.960 & 0.870 & 0.470 & 0.670 \\
\midrule
SDXL-Turbo      
& 0.840 & 0.200 & 0.090 & \textbf{1.000} & 0.660 & 0.450 & 0.540 \\
\rowcolor{lightgray}
+ ReNO          
& 0.900 & 0.350 & 0.130 & \textbf{1.000} & 0.840 & 0.680 & 0.650 \\
\midrule
HyperSDXL       
& 0.870 & 0.210 & 0.100 & \textbf{1.000} & 0.760 & 0.430 & 0.560 \\
\rowcolor{lightgray}
+ ReNO          
& \textbf{0.910} & 0.330 & 0.170 & \textbf{1.000} & 0.900 & 0.560 & 0.650 \\
\midrule
Flux-Schnell    
& 0.780 & 0.430 & 0.180 & 0.980 & 0.800 & 0.640 & 0.640 \\
\rowcolor{lightgray}
+ ReNO          
& 0.870 & 0.560 & 0.210 & 0.990 & 0.900 & \textbf{0.790} & 0.720 \\
\bottomrule
\end{tabular}}
\end{table}

\end{document}